\documentclass{article}

\usepackage{arxiv}

\usepackage[utf8]{inputenc} 
\usepackage[T1]{fontenc}    
\usepackage{hyperref}       
\usepackage{url}            
\usepackage{booktabs}       
\usepackage{amsfonts}       
\usepackage{nicefrac}       
\usepackage{microtype}      
\usepackage{lipsum}
\usepackage{graphicx}
\graphicspath{ {./images/} }

\title{Analysis of Model-Free Reinforcement Learning Control Schemes on self-balancing Wheeled Extendable System}

\author{
 Kanishk . \\
  Department of Applied Physics\\
  Delhi Technological University\\
  New Delhi, India \\
  \texttt{k.kanishk@gmail.com} \\
   \And
 Rushil Kumar\\
  Department of Applied Physics\\
  Delhi Technological University\\
  New Delhi, India \\
  \texttt{rushilkumar18@gmail.com} \\
  \And
 Vikas Rastogi\\
  Department of Mechanical Engineering\\
  Delhi Technological University\\
  New Delhi, India \\
  \texttt{vikasrastogi@dtu.ac.in} \\
  \And
 Ajeet Kumar\\
  Department of Applied Physics\\
  Delhi Technological University\\
  New Delhi, India \\
  \texttt{ajeetdph@dtu.ac.in} \\
}

\begin{document}
\maketitle
\begin{abstract}
Traditional linear control strategies have been extensively researched and utilized in many robotic and industrial applications and yet they don’t respond to the total dynamics of the systems. To avoid tedious calculations for nonlinear control schemes like H-infinity control and predictive control, the application of Reinforcement Learning(RL) can provide alternative solutions. This article presents the implementation of  RL control with Deep Deterministic Policy Gradient and Proximal Policy Optimization on a mobile self-balancing Extendable Wheeled Inverted Pendulum (E-WIP) system with provided state history to attain improved control. Such RL models make the task of finding satisfactory control schemes easier and responding to the dynamics effectively while self-tuning the parameters to provide better control. In this article, RL-based controllers are pitted against an MPC controller to evaluate the performance on the basis of state variables and trajectory errors of the E-WIP system while following a specific desired trajectory.
\end{abstract}


\section{Introduction}
Control  systems have come a long way in the past few decades. Feedback control is really popular in academic research and the industry. SISO feedback proves to be quite efficient for simple linear systems with directly dependent observed quantities and input. Usually, for controlling more complex nonlinear systems, linear approximation around the desired control output is calculated and linear control systems are designed around those linear estimations. MIMO feedback control is extremely useful in many cases as it not only incorporates various observable quantities but also different controllable quantities as well. Many MIMO architectures have been developed over the years, and these control strategies employ modern control methods like predictive control and the application of machine learning in control systems.
\begin{figure}[bt]
\centering
\includegraphics[width=10cm]{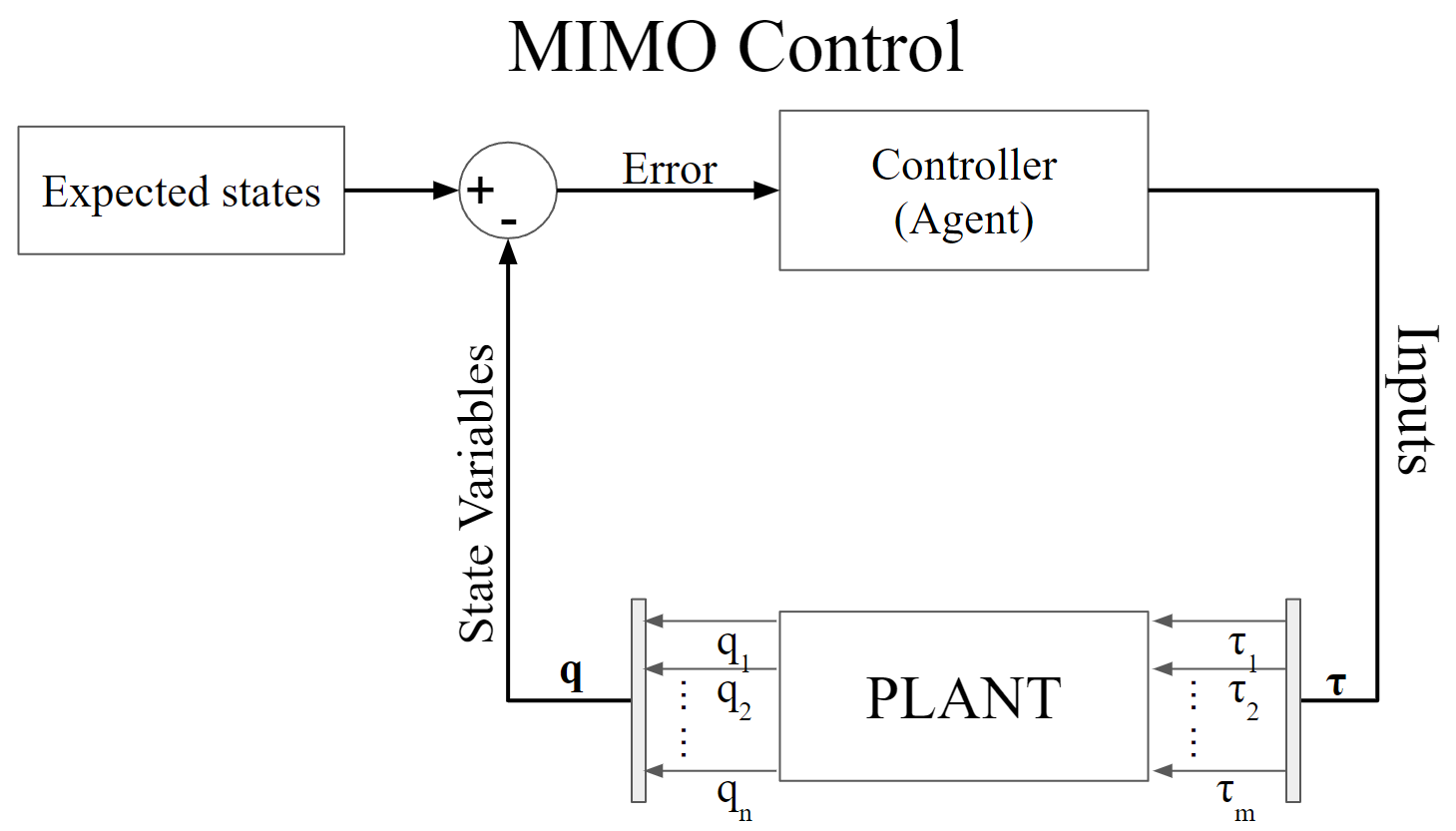}
\caption{Basic Multi In Multi Out control architecture.}
\end{figure}
The Inverted pendulum on a cart is an ideal unstable system problem thereby making it one of the classic problems revolving around feedback control theory, control systems and control algorithms. A load is attached to the top of a wheeled assembly that moves along a track. One of the earliest researches on integrating neural networks in order to come up with a control system for the Inverted pendulum on a cart was by Charles W. Anderson. He described two layered and two single layered neural network system using a rudimentary form of reinforcement learning and temporal difference technique to solve the given control problem. The paper mentions only one state of measurement which is the number of tasks ‘failures’ and also indicates the tedious amount learning time required \cite{Anderson1989-cg}. 

Moving into the recent studies of similar configuration systems using different control systems such as PD, LQR, Fuzzy Logic, MPC and RL controllers. The comparison of all these algorithms and control systems are important in terms of robustness, timeliness, integration, acceptability and flexibility in order to come up with the best solution for industry application, authors reviewed papers comparing control systems mentioned above, in the same (or similar) assembly; such as comparison of PID vs (PD like) Fuzzy logic control in a similar configuration done by Goher\cite{Goher2019-yv} in 2019, concluded that the (PD like) Fuzzy logic control provides much better performance than PID controllers improving on the system responses. The work on Model Predictive Control for a Wheeled Inverted Pendulum by Yue et. al. \cite{Yue2018-js} shows the excellent robustness of MPC controller in the WIP system. The study performed by Dinev, T. et. al. \cite{Dinev2020-er} on a wheeled robot had a prismatic joint as an extensible architecture that allowed an extra DOF in the direction of translation of the joint. They further compared MPC and PD controllers on their system in order to benchmark the qualities of the controllers, showing that MPC and PD showed identical results on flat terrain. However, on rough terrain MPC had better stabilization, and robust performance in relation to sensor noise. MPC seems to have a very robust and less variable performance on the inverted pendulum system. While comparing LQR to PD control, LQR proves to be a better alternative for PD controllers as concluded in the results by Li et al. \cite{Li2014-qh}. LQR proves to be efficient in tracking small positional change which makes the system more responsive and improves the overall performance. Although, as discussed in this paper \cite{Li2014-qh} nonlinear control systems are necessary in order to achieve more manageable solution for larger displacement.

Reinforcement learning based controllers in recent years have shown essential progress in understanding non-linear control problems, and variable performance in learning time and robustness have also seen ample improvement in recent works based on the Reinforcement learning based controllers \cite{Li2014-qh}, \cite{Dao2021-oe} and Lim et al. \cite{Lim2020-el}. Getting our inspiration from the work being done in the field of applying RL in mobile robot control, we have developed an Extendable Wheeled Inverse Pendulum (E-WIP) system to control with two different popular RL algorithms.

Wheeled Inverse Pendulum systems are being extensively investigated for their applications in modern robotics for creating mobile robots. Kim et al. \cite{Kim2005-id} analyzed and demonstrated a two wheeled IP using LQR based controller, for basic operation like balancing, steering and spinning. They claimed that the robustness of the model they presented was on-par with their expectations and they proved it by testing their system on different terrains. Peng et al.\cite{peng2020-et} stated the use of PID along with Fuzzy Logic Controllers to balance a modified Two-Wheeled Inverted Pendulum robot. They introduced a prismatic joint with a bar at the top of system for better balance and then designed Fuzzy Logic Controller for this system. Similar to Wang et. al, many modified versions of inverse pendulum have been studied over time with new and advanced control strategies for accomplishing various tasks. Works like that of Klemm et al., \cite{Klemm2019-cy} and Dinev et al., \cite{Dinev2020-er} shows the advent of modified Inverse pendulum systems in the hopes of better balancing performance. The former mentions the use of LQR controllers and later compares the model predictive controllers to traditional controllers like PD for crossing rough terrains. In recent times, reinforcement learning has been a major field of interest for system design researchers. Manrique Escobar et al., \cite{Manrique_Escobar2020-tt} shows the implementation of deep reinforcement learning in controlling cart-pole system. Such works have motivated the authors to experiment with new RL-based control strategies and develop controllers for the E-WIP system due to it versatile and nonlinear nature. In this article, the authors discuss Deep Reinforcement and model predictive strategies and error propagation in policy networks to control a simulated Extendable wheeled inverted pendulum system.
\section{System Definition}
The control system analysis is done on an Extendable wheeled inverted pendulum(E-WIP). The independent states of the system can be represented by $\dot{q}=\left[\begin{array}{lllll} x & z & \theta & \phi & l\end{array}\right]^{T}$ and their derivatives are given by $\dot{q}=\left[\begin{array}{lllll}\dot{x} & \dot{z} & \dot{\theta} & \dot{\phi} & \dot{l}\end{array}\right]^{T}$, where $x$ is the displacement of wheel in x direction and $z$ is the displacement in z axis, $\theta$ is the angle made by the pendulum with the normal and  $\phi$ is the angle the wheel has been rotated and finally, $l$ is the length of the Extendable pendulum from the wheel center to the ‘bob’.
\begin{figure}
\centering
\includegraphics[width=10cm]{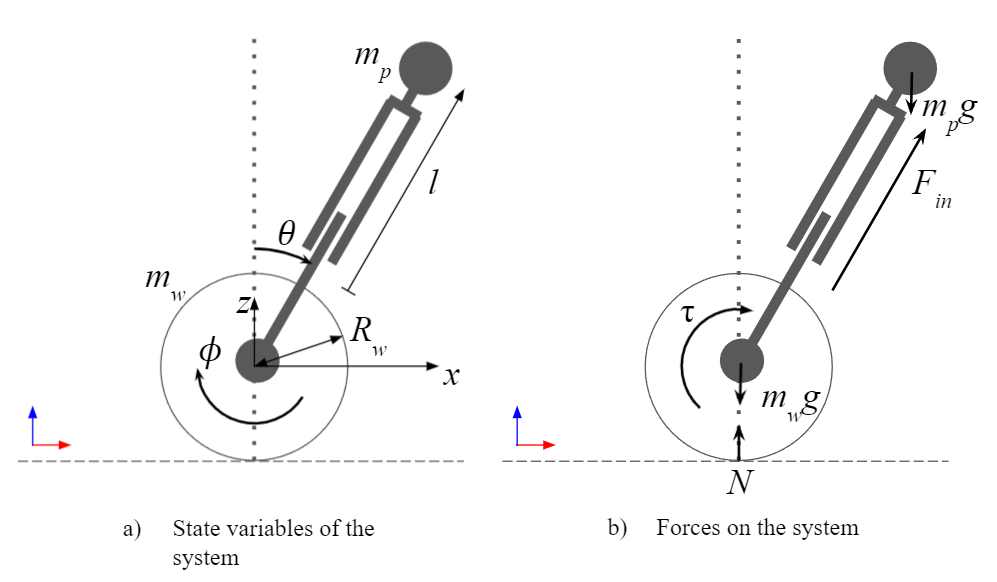}
\caption{Extendable Wheeled Inverse Pendulum System state variables and applied forces on the system}
\end{figure}
The non-linear equations of motions of this system can be written by writing the total kinetic ($T$) and potential energy ($V$) of the system. They are given by 
\begin{equation}
    T=\frac{1}{2} m_{w}\left(\dot{x}^{2}+\dot{z}^{2}\right)+\frac{1}{2} m_{p}\left(\dot{x_{p}}^{2}+\dot{z_{p}}^{2}\right)+\frac{1}{2} I \dot{\phi}^{2}
\end{equation}
\begin{equation}
    V=m_{w} g z+m_{p} g z_{p}
\end{equation}
Where $x_p$ and $z_p$ are defined for the pendulum ‘bob’ as,

\begin{equation}
    x_{p}=x+l \sin (\theta)
\end{equation}
\begin{equation}
    z_{p}=z+l \cos (\theta)
\end{equation}

Here, $m_w$ is the mass of the wheel, $m_p$ is the mass of the pendulum ‘bob’, and g is the acceleration due to gravity. Let the Lagrangian be $\mathcal{L}=T-V$, for the various state variables, the state equations in terms of Lagrangian can be written as,
\begin{equation}
    \frac{d}{d t} \frac{\partial \mathcal{L}}{\partial \dot{x}}-\frac{\partial \mathcal{L}}{\partial \mathrm{x}}=\frac{\tau R_{w}}{I}+F_{i n} \sin (\theta)
\end{equation}
\begin{equation}
    \frac{d}{d t} \frac{\partial \mathcal{L}}{\partial \dot{z}}-\frac{\partial \mathcal{L}}{\partial \mathrm{z}}=m_{t} g+F_{i n} \cos (\theta)+N
\end{equation}
\begin{equation}
    \frac{d}{d t} \frac{\partial \mathcal{L}}{\partial \dot{\theta}}-\frac{\partial \mathcal{L}}{\partial \theta}=-\tau
\end{equation}
\begin{equation}
    \frac{d}{d t} \frac{\partial \mathcal{L}}{\partial \dot{\phi}}-\frac{\partial \mathcal{L}}{\partial \phi}=-\tau
\end{equation}
\begin{equation}
    \frac{d}{d t} \frac{\partial \mathcal{L}}{\partial \dot{l}}-\frac{\partial \mathcal{L}}{\partial l}= F_{in}
\end{equation}

In Equations 5-9, $ N $ is the normal force on the wheel, $ \tau$ is the input torque on the wheel by the motor and $ F_{in} $ is the input force on the Extendable link. $ \Ddot{q}$ can be written separately for each state variable to define the nonlinear dynamics of the system.

\begin{equation}
    \frac{d^{2} x}{d t^{2}}=\frac{\tau\left(I \cos (\theta)+R_{w} l\right)}{I m_{w} l}
\end{equation}
\begin{equation}
    \frac{d^{2} z}{d t^{2}}=\frac{N}{m_{w}}-\frac{g m_{p}}{m_{w}}-\frac{\tau \sin (\theta)}{m_{w} l}-2 g
\end{equation}
\begin{equation}
    \frac{d^{2} \theta}{d t^{2}}=\frac{-I m_{p} m_{w}(g \sin (\theta)+2 \dot{l} \dot{\theta}) l+I m_{p}\left(-g m_{p}+N\right) l \sin (\theta)-I m_{p} \tau-I m_{w} \tau-R_{w} m_{p} l \tau \cos (\theta)}{I m_{p} m_{w} l^{2}}
\end{equation}
\begin{equation}
    \frac{d^{2} \phi}{d t^{2}}=\frac{\tau}{I}
\end{equation}
\begin{equation}
    \frac{d^{2} l}{d t^{2}}=g \frac{m_{p} \cos (\theta)}{m_{w}}+g \cos (\theta)+l \dot{\theta}^{2}-\frac{N \cos (\theta)}{m_{w}}+F_{i n}-R_{w} \tau \frac{\sin (\theta)}{I m_{w}}
\end{equation}

The above equations (10-14) are highly non-linear in nature, so to design optimal control strategy, a complex controller needs to be defined by more rigorous calculations involving inseparable nonlinear terms. In order to bypass the calculations, a self-learning control system can be implemented that moves ‘closer’ towards the ‘ideal’ controller for this system through iterative training.

\section{Control Schemes}
\subsection{Reinforcement Learning in Control System Design}
Controlling and handling highly complex systems such as; mechanical assemblies with an extensive amount of dynamics involved \cite{Villecco2017-zl}, with the traditional approaches of deducing the system equations to compute their inverse or forward kinematics and tedious adjustment of parameters is rather unyielding and a gargantuan task. A more manageable solution to such highly dynamic and complex problems can be attained with the help of ML systems and algorithms. One such ML algorithm is based upon policy iteration of Markov Decision Process (MDPs) known as Reinforcement Learning. Another thread of RL was based upon trial and error \cite{Sutton2018-ze}, much analogous to how humans psychologically learn a task, via the reward-punishment based skill development, where reward is given at a successful execution and punishment when a task fails \cite{Pezzulo2007-ci} \cite{Xiang2019-vh}. Figure 3 shows a typical RL-based controller attached to a plant.
\begin{figure}
\centering
\includegraphics[width=10cm]{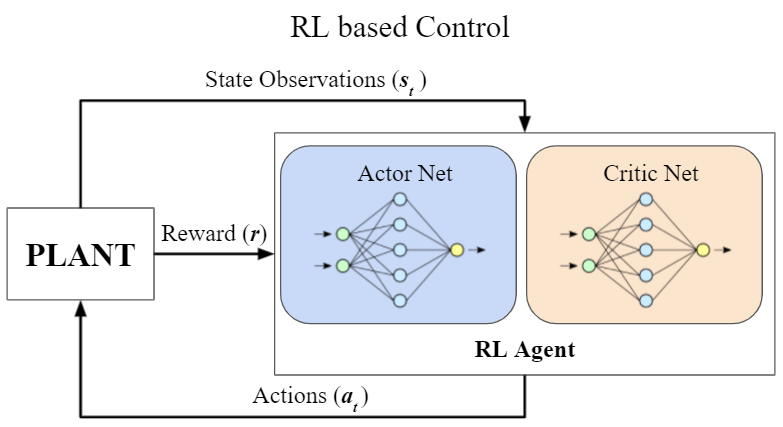}
\caption{Typical RL-based control setup.}
\end{figure}
This MDP optimizing technique has a certain goal of maximizing its reward functions, considering a uniform and standard RL setup which has an agent (consisting of an actor and a critic), this given agent interacts with the Environment in given specific discrete time. The parameters involved in agent-environment interaction are ($S_t,A_t,S_{t+1},R_{t+1}$). $S_t$ gives state of the environment at time ‘t’, At gives the action performed by the agent at time ‘t’, $S_{t+1}$ describes the state of the environment after action $A_t$ has taken place at time ‘$t+1$’, $R_{t+1}$ describes the reward received by the agent after evaluation of $S_{t+1}$. As mentioned earlier the goal of the system is to imply an ‘optimal control’ policy $\pi:\ \ \ S_t\rightarrow A_t$ (Which maps the states over actions) with maximized amount of rewards amassed at each time step \cite{Zhang2018-qj}, i.e. total cumulative reward to be maximized can be expressed as:
\begin{equation}
    R_{total}=\ \sum_{t}^{T_\infty}{R_t\ \ \ }
\end{equation}
The policy $\pi$ can be approximated as a deep neural network function which implies that it contains weights of the various neurons and layers given by a vector $\vartheta$. Now, $\pi_{(s,a,\vartheta)}$ becomes a deep neural network function that receives the states, actions and the weights to give the control policy.
The validity of any RL based controller is defined by three basic functions as described by Zhang et al. the appropriate state value function $\mu_\pi(s)$, the appropriate action value function $q_\pi(s,a)$ and the appropriate action value function w.r.t $a$ state $a_\pi(s,a)$. For current state $S_t=s$, current action $A_t=a$, time $t$, and iteration $k$, these functions on an environmental expectation $\mathbb{E}$ with the action policy $\pi$ become,
\begin{equation}
v_{\pi}(s) \doteq \mathbb{E}_{\pi_{s, \vartheta}}\left[\gamma^{k} R_{t+k+1} \mid S_{t}=s\right]
\end{equation}
\begin{equation}
q_{\pi}(s, a) \doteq \mathbb{E}_{\pi_{s, a, \vartheta}}\left[\gamma^{k} R_{t+k+1} \mid S_{t}=s \quad A_{t}=a\right]
\end{equation}
\begin{equation}
a_{\pi}(s, a) \doteq q_{\pi}(s, a)-v_{\pi}(s)
\end{equation}

Here $\gamma$ is the discount factor. The goal of training the RL network is to learn the $\vartheta$ vector to find the optimal control policy $\pi$.
In recent developments of RL, research has come up with numerous algorithms primarily for state continuity and action spaces, out of them deep deterministic policy gradient(DDPG)[18,21] and proximal policy optimization(PPO)\cite{SchulmanWDRK17} are the policy gradient methods of RL implementations, we will be benchmarking for our analysis. These methods have iterated optimisation of  policy estimated by the gradients. DDPG and PPO are both model-free algorithms, these algorithms follow the actor-critic architecture, their goal is to maximize the long-term reward, reducing the deviation of the gradient estimate.\cite{Bohn2019-ta} \cite{lillicrap2019continuous}.
\subsubsection{Deep Deterministic Policy Gradient}
Deep Deterministic Policy Gradient (or DDPG) is a model-free algorithm actor-critic framework while learning a deterministic policy expanding the work to continuous space rather than discrete spaces via the combination of DPG and DQN [18,21]. The DDPG algorithm trains two networks, the actor $\mu_\vartheta(s)$ and the critic $Q_\varphi(s,a)$. As mentioned earlier, $\mu_\vartheta(s)$ gives the actual actions corresponding to a given state and $Q_\varphi(s,a)$ is the Q-value which measures the ‘goodness’ of the action taken for the current state.
The neural networks this defined are randomly initializes and are then subjected to mini batches from the experience buffer. For each experience, sets of state variables, actions, reward and the state after the action is pushed to update the actor and critic networks to take the appropriate action for the given state and get the optimal Q-value respectively. The critic or the Q-value network is updated by minimizing the Mean squared Bellman error as described by Escobar et al. \cite{Manrique_Escobar2020-tt}
\begin{equation}
    L_{B}=\frac{1}{M} \sum_{i=1}^{M}\left(Q_{\varphi}\left(s_{i}, a_{i}\right)-y_{i}\left(s_{i}, a_{i}, r_{i}, s_{i}^{\prime}\right)\right)^{2}
\end{equation}
Here M is the mini batch size, $r_i$ is the current reward and $y_i$ is the expected target value which can be expressed as,
\begin{equation}
    y_{i}=\left\{\begin{array}{c}
r_{i}+\gamma Q_{\varphi}^{\prime}\left(s_{i}^{\prime}, \mu^{\prime}\left(s_{i}^{\prime}\right)\right), t<t_{\text {end }} \\
r_{i}, t=t_{\text {end }}
\end{array}\right.
\end{equation}
Here, $\gamma$ is the discount factor.
$L_B$ is utilized to update the Q network’s parameters $\varphi$ and the gradient $\nabla$ is utilized to update the policy or the actor $\mu$’s parameters $\vartheta$ by keeping parameters $\varphi$ of the $Q$ network constant. The policy gradient is given by differentiating $Q_\varphi$ w.r.t $\vartheta$ in $pi$,
\begin{equation}
    \nabla=\frac{1}{\mathrm{M}} \sum_{i=1}^{M} \nabla_{\mu_{\vartheta}\left(s_{i}\right)} Q_{\varphi}\left(s_{i}, \mu_{\vartheta}\left(s_{i}\right)\right) \nabla_{\vartheta} \mu_{\vartheta}\left(s_{i}\right)
\end{equation}
The policy and the value function are updated until a viable policy is found.
In our case, the state and the action can be described in tuples of the following format,
\begin{equation}
s=(x, z, \theta, \phi, l, \dot{x} \quad \dot{z} \quad \dot{\theta} \quad \dot{\phi} \quad i)
\end{equation}
\begin{equation}
a=\left(\tau, F_{\text {in }}\right)
\end{equation}
\begin{figure}[bt]
\centering
\includegraphics[width=12cm]{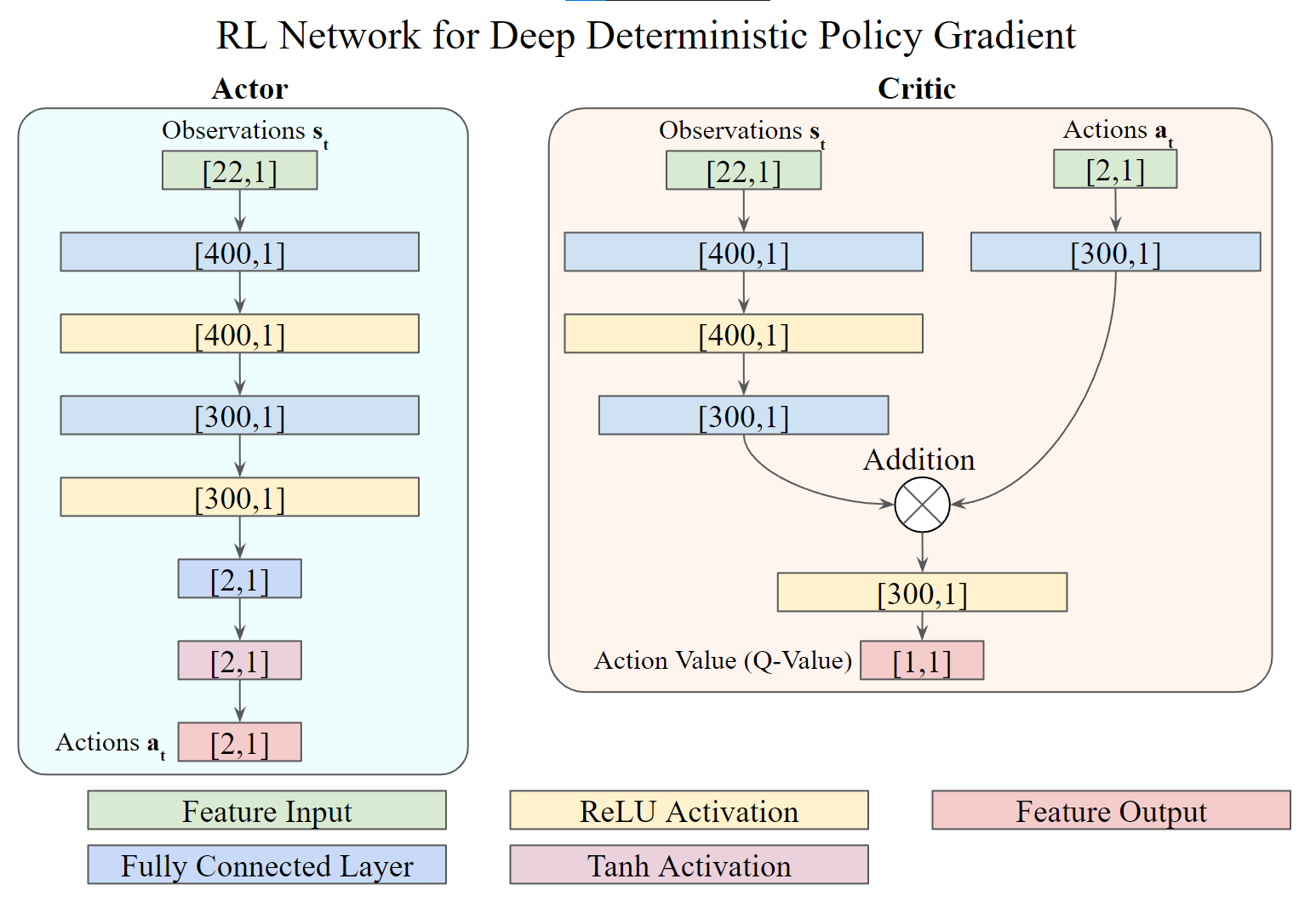}
\caption{RL network architecture for DDPG Agent}
\end{figure}

\subsubsection{Proximal Policy Optimization}
PPO (proximal policy optimization) is defined as a model-free policy gradient reinforcement learning technique that aims to evaluate or improve the policy that is used to make choices ('On Policy'). This technique is a form of policy gradient training that alternates between sampling data via interaction with the Environment and optimizing a section of the objective function using stochastic gradient descent. The clipped surrogate goal function improves training stability by limiting the extent of policy changes at each step \cite{SchulmanWDRK17}. Unlike DDPG, PPO calculates its objective function by getting the ratio of the next and the previous policy and subjecting that to an advantage function $\hat{A}$ . The surrogate objective function is given as,
\begin{equation}
    L(\vartheta)=\widehat{\mathbb{E}}_{t}\left[\min \left(r_{t}(\vartheta) \hat{A}_{t}, \operatorname{clip}\left(r_{t}(\vartheta), 1-\epsilon, 1+\epsilon\right) \hat{A}_{t}\right)\right]
\end{equation}
Here, $r_t(\vartheta)$ denotes the ratio of the new and old probability distributions of $\mu(a_t,s_t)$ also known as the divergence of the distributions. The critic depicts a state value function $V_t(s_t)$ which is utilized to define the ‘advantage’ $\hat{A}_t$ for the current time step.
\begin{equation}
    \widehat{A}_{t}=-V\left(s_{t}\right)+r_{t}+\gamma r_{t+1}+\ldots+\gamma^{T-t+1} r_{T-1}+\gamma^{T-t} V\left(s_{T}\right)
\end{equation}
$T$ is the maximum time steps forward the value estimator can foresee which adds to the overall advantage $\hat{A}_t$t. For each epoch, the actor calculated the surrogate losses for each time segment of length $T$. The advantage is calculated and then we optimize the parameters $\vartheta$ for a minibatch of $M$.
Similar to DDPG, the state variables and actions for the system presented in this paper are,
\begin{equation}
s=(x, z, \theta, \phi, l, \dot{x} \quad \dot{z} \quad \dot{\theta} \quad \dot{\phi} \quad i)
\end{equation}
\begin{equation}
a=\left(\tau, F_{\text {in }}\right)
\end{equation}
\begin{figure}[bt]
\centering
\includegraphics[width=12cm]{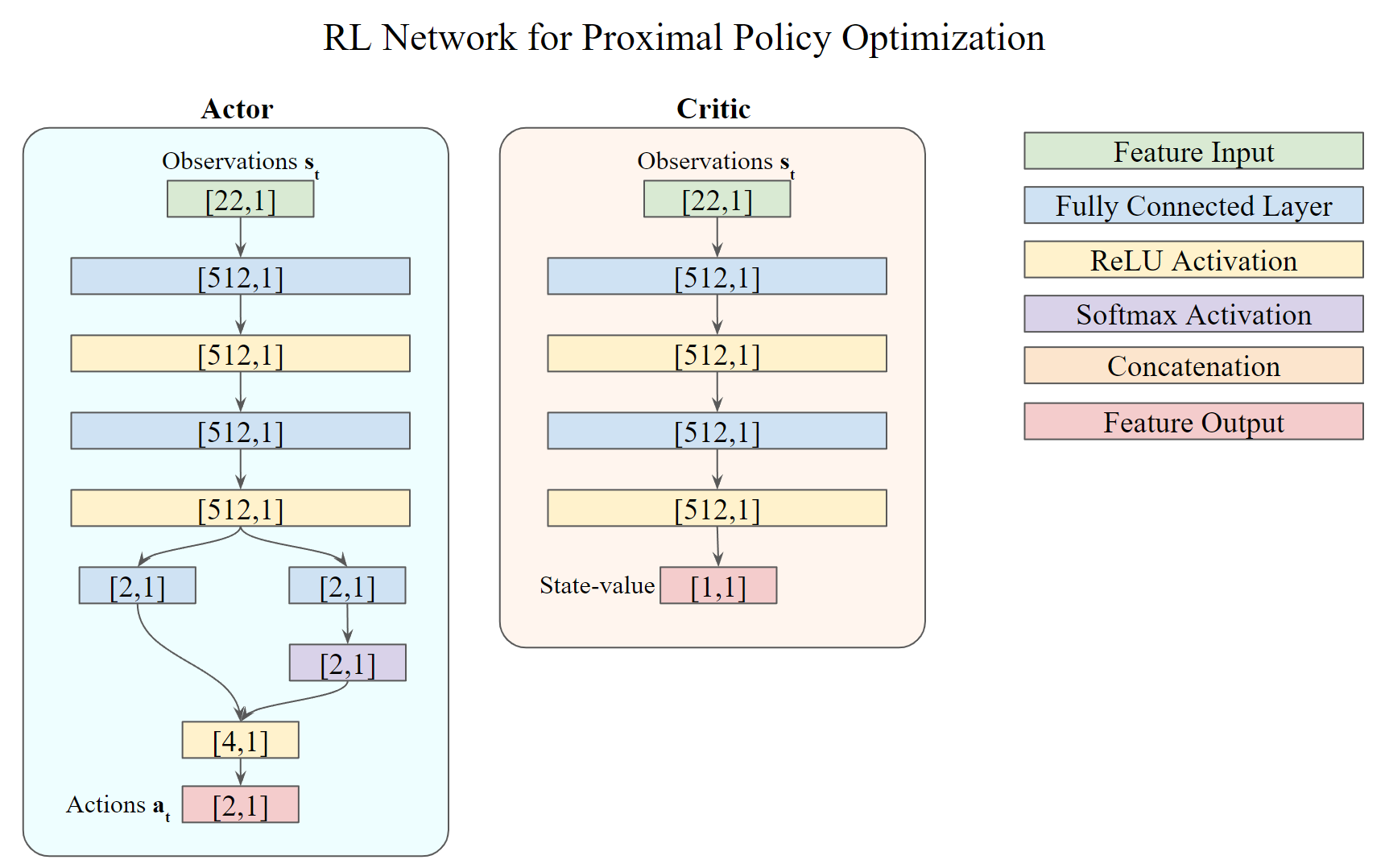}
\caption{RL network architecture for PPO Agent}
\end{figure}

\subsection{Error History Dependent implementation of DDPG and PPO}
To make our RL models perform better, previous error states (Error History (EH)) have been provided as observations to the agents. The error consists of expected and current state values, here x and z. The errors of previous 6 time-steps have been included in the observations. The total number of observations become 22 (10 state variables $q$ and $\dot{q}$, 6 previous error pairs for x and z). The observation vector becomes,
\begin{multline}
    s_t=(x, z, \theta, \phi, l, \dot{x}, \dot{z}, \dot{\theta}, \dot{\phi}, \dot{l},e_{1,x}, e_{1,z}, e_{2,x}, e_{2,z}, e_{3,x}, e_{3,z}, e_{4,x}, e_{4,z}, e_{5,x}, e_{5,z}, e_{6,x}, e_{6,z})
\end{multline}
Here $e_{n,x}$ is the nth timestep delayed error in the x value from the reference x trajectory. The action vector can be written as $a=\left(\tau, F_{\text {in }}\right)$. The main idea behind propagating the errors in the states with reference is to keep the relation between the reference states and the current states in the loop of updating the policy. The target is to find a better optimal control policy that follows the reference state variables given in form of a trajectory.

The RL environment for DDPG and PPO has the update time step of 0.05 and 0.01 seconds respectively. The experience buffers for both techniques have been set to the max size of 1e6, with discount factors of 0.99. The learning rates of DDPG's actor and critic are 1e-5 and 1e-4, whereas the PPO's actor and critic have learning rates of 1e-4. Both techniques utilizes Adam optimizer for updating the networks. 

\subsection{Model Predictive Control}
Model predictive control (MPC) is an effective and efficient internal control approach which has been in academic and research discussions. It is broadly applied in industrial operations. MPC's performance tuning formulation and ability to handle restricted and nonlinear multivariable problems distinguish it from traditional feedback control methods. However, as compared to conventional proportional-integral-derivative (PID) controllers, MPC's computing efficiency seems to be at a disadvantage, especially when used in large-scale nonlinear programming (NLP) challenges; as NLP must be handled online at every time interval \cite{He2020-pq}.
Therefore, using MPC for NLP requires the use of methods such as Sequential Quadratic Programming. The SQP technique is an iterative approach based on actively improving the system that solves a series of optimization subtasks using warm-start and optimal active-set recognition \cite{Kouvaritakis2016-fx}. The linear MPC employed here creates a Hessian matrix which contains the prediction model based on the number of inputs, the observable state, added input and output noise and the prediction horizon. This prediction matrix is viewed as an optimization QP problem is solved by using the Active-set solver.

The MPC controller implemented here is based on a linearized model of the system around its desired position where the system has no velocity and stands still with $\theta=0$. The MPC is created using the MPC designer tool box in Matlab. Sampling rate is 0.01s with prediction horizon of 100 and control horizon of 15. The input observations are the 10 state variables and the two outputs, $\tau$ and $F_{in}$ respectively. Since we have the reference values for $x$ and $z$, and the MPC has the constraint of keeping the $\theta$ value between $-\pi/6$ and $-\pi/6$. Other states are unconstrained. The weights on input variables $\tau$ and $F_{in}$ are 0.0210 and 0.2101 respectively and the rate of change of input variables i.e. $\tau$ and $F_{in}$ have the weight of 0.4759 each.

Since MPC controllers are fairly popular and provide robust system control in many situations, here, the comparison of the RL models developed is done against the MPC for the same system and also for the same input trajectories.

\section{Experimental Setup}
The Simulation is developed and run on Simulink. This paper focuses on the ability of the robot system under study to balance itself, move and stop at a point. The system comprises of the Robot and a ground plane which provides friction for the robot wheel. The Robot itself is made up of three major parts, the wheel, the Extendable link and the bob. The dimensions of the robot’s anatomy are given in the figure. The wheel is connected to the Extendable link with a revolute joint and the bob is connected to the wheel shaft with a prismatic joint. Other simulation parameters are expressed in the Table 1.

\begin{table}
\centering
\caption{System parameters}
\begin{tabular}{l c c}
\toprule
Quantity & Value & Unit\\
\midrule
Coefficient of Friction of The Ground&	0.6&	-\\
Radius of Wheel&	100&	mm\\
Mass of Wheel&	0.25&	kg\\
Link Minimum Length&	250&	mm\\
Link Maximum Length&	500&	Mm\\
Radius of Bob&	50&	mm\\
Mass of Bob&	0.125&	kg\\
\hline  
\end{tabular}
\end{table}

\begin{figure}[bt]
\centering
\includegraphics[width=12cm]{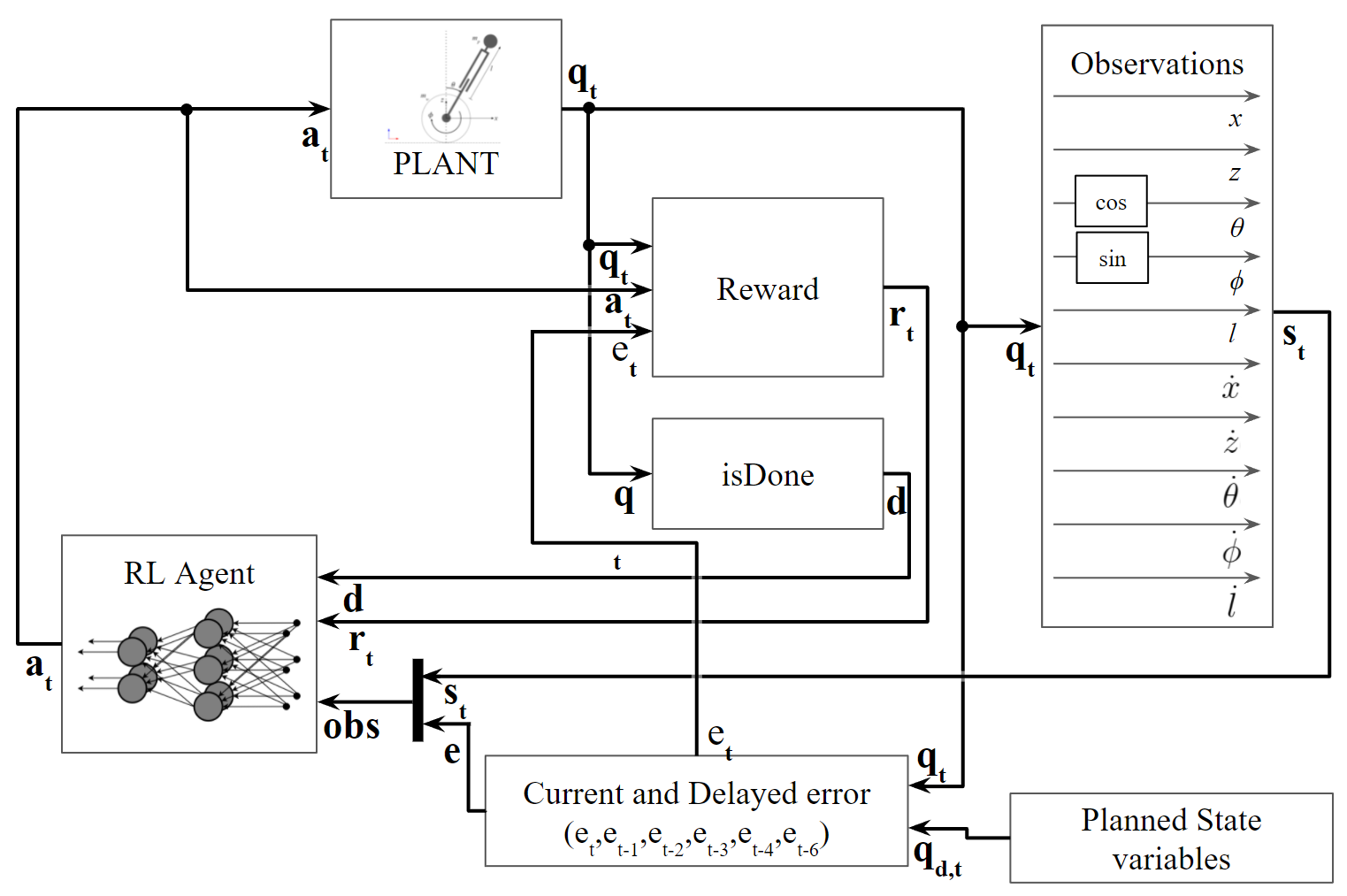}
\caption{Detailed Simulink Setup}
\end{figure}

The RL simulation is also setup in Simulink (Figure 6). The raw states generated from the environment are processed to from a part of observations ($\boldsymbol{s_t}$) to be fed to the RL block. Planned variables contains the $x$ and $z$ position of the trajectory for the system to follow. The errors are calculated from the current system’s $x$ and $z$ positions and the errors of last 5 time-steps ($\boldsymbol{e}$) from the total Observation vector. The RL block also requires a ‘done’ signal $d$ to terminate and reset the simulation when required.
\begin{equation}
    \boldsymbol{d}=\left\{\begin{array}{c}
0, \text { otherwise } \\
1, \text { if }\left(\theta \leq \frac{\pi}{4}\right) \text { and }(|x| \leq 4)
\end{array}\right.
\end{equation}

The reward ($r$) depends on the actions, state variables, previous $x$ and $z$ change rates and the trajectory error. The reward function for this simulation can be written as,
\begin{equation}
    \boldsymbol{r}_{t}=r_{f}+r_{v}+0.3
\end{equation}
\begin{equation}
    r_{f}=-\left\{0.2 \theta_{t}^{2}+0.25 \dot{\theta_{t}}^{2}+0.02\left(\tau_{t}+F_{i n_{t}}\right)+0.75 e_{x, t}+0.05 e_{z, t}+0.5\left(\dot{x}_{t-1} +\dot{z}_{t-1}\right)\right\}
\end{equation}
\begin{equation}
    r_{v}=\left\{\begin{array}{c}
        0, \text { otherwise } \\
        -50, \text { if }\left(\theta \leq \frac{\pi}{4}\right) \\
        -100, \text { if }(|x|<4)
\end{array}\right.
\end{equation}

\section{Result and Discussion}
The training of the proposed ML models is done on a cloud VM with an Intel Xenon 2.3GHz 4 core CPU, 52 GB memory and a Nvidia Tesla T4 GPU with 16GB of compute memory. The training for both models was done in discrete manner due to the memory constraints of the system. Since the system is continuous, the training took a considerable amount of time. The actor and critic are trained in parallel. 

For experimentations, we compared traditional DDPG and PPO algorithms with 10 states as input observations with our modified Error History based DDPG and PPO along with an optimal MPC controller. Table 2 depicts the training parameters for the actor and critic agents for all test models. Since we are studying the balancing and the point-to-point motion of the robot in one direction, we ‘publish’ the trajectory values to the robot. The ‘ideal’ trajectory requires the robot to stay at its position for 3 seconds, then reach 2 meters ahead in the next 4 seconds and then stay at the destination for 3 more seconds. MPC performs well but possesses a delay throughout the trajectory but does the best work at stabilizing the system when it stops. DDPG controllers do the best job at following the exact trajectory given to it. The DDPG without error history does a worse job at stopping in comparison to the error history based DDPG which was the best RL controller amongst all. It even provides good stabilization when the system is at rest. On the other hand, PPO controllers do provide stability in the start, does a decent job at following the trajectory but they do a relatively bad job at slowing down in the last 3 seconds especially the PPO controller without error history. This can be observed in Figure 7.

\begin{table}
\centering
\caption{Experiment Parameters}
\begin{tabular}{lcccc}
\toprule
\bf{Parameter} & \bf{DDPG$^{EH}$} & \bf{DDPG} & \bf{PPO$^{EH}$} & \bf{PPO}\\
\midrule
\bf{Sample Time} &	0.05  & 0.05 &	0.01 & 0.01\\
\bf{Episodes} & 8578 & 9457 &	9875 & 10015\\
\bf{Mean Max Reward} &	55.47 & 49.75 &	235.12 & 175.48\\
\hline  
{$^{EH}$Error History based}
\end{tabular}
\end{table}


\begin{figure}[!t]
\centering
\includegraphics[width=14cm]{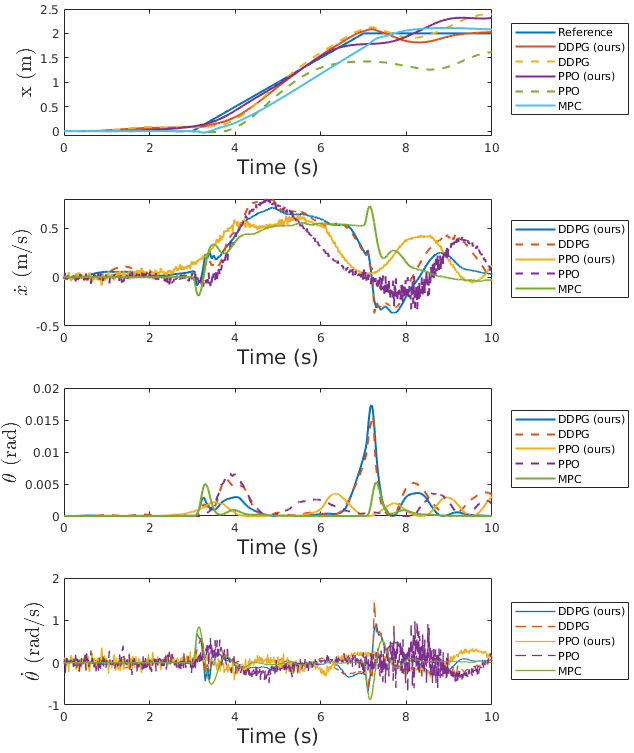}
\caption{State variable comparison for DDPG, PPO and MPC based control schemes.}
\end{figure}
\begin{figure}
\centering
\includegraphics[width=14cm]{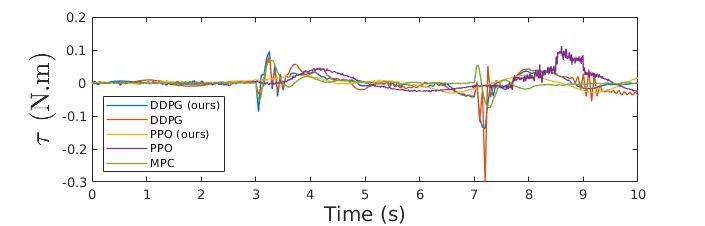}
\caption{Torque input comparison for DDPG, PPO and MPC based control schemes}
\includegraphics[width=14cm]{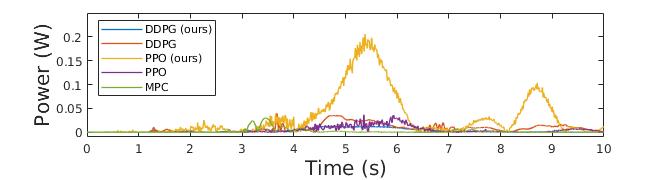}
\caption{Rotational Power comparison for DDPG, PPO and MPC based control schemes}
\end{figure}

The positional data in Figure 7 shows that all control strategies are successful in following the reference trajectoty. MPC seems to fall behing while following the path while RL algorithms try to accurately follow the given positional waypoints. From the velocity curves in Figure 8, it is evident than DDPG with EH and MPC tend to go hand in hand with each other but only with fluctuations during stopping. Both PPOs directly starts to increase its velocity without accelerating backwards slightly to build up forward momentum like MPC and DDPG controllers but slows don in the middle to compensate for the excessive tilt. All controllers achieve a maximum X velocity of 0.5 m/s. There is a much larger angular deviation at the stopping point for the DDPG with EH controller, but it stabilizes this deviation quickly and returns back to 0. The PPO Controller shows a great amount of high frequency angular velocity fluctuations, this indicate that the controller needs more time and episodes to train to smooth out the noise.

Figure 8 shows the input torque provided by the controller to the system. DDPG with EH controller moves closely with the MPC but with disturbances at some points. PPO’s input contain significant noise and it follows a different control strategy than the MPC and DDPG. Its effects are clearly visible in the state outputs. Rotational power of the wheel can be expressed as, $P = \tau \dot{\phi}$, where $\dot{\phi}$ is the wheel's angular velocity and $\tau$ is the input torque on the wheel. The rotational power of the wheel under RL algorithms for the same trajectory is shown in the following Figure 9. PPO controller pumps in the most energy into the wheel, indicating that this control scheme might be very inefficient w.r.t DDPG and MPC.Our implementation of DDPG is on-par with the predictive MPC controller. DDPG based controller not only proves to be good in following the given trajectory, but also the wheel possesses less rotational power even less than MPC; this is due to the introduction of input penalty introduced in the reward function (Equation 31). The only limitation is seen in PPO controller. A suspected reason is the lack of experience for the stopping portion of the task training process. Even for greater training episodes and steps, PPO does not seem to perform as well as the other two controllers. This may be due to the fact that due to limited computational resources, PPO network might have not been trained sufficiently enough to remove the noise form its inputs. PPO seems to work better than DDPG on paper but this is not seen in our experiments, the only way to improve PPO’s results is to perform more rigorous training on more computationally powerful machines with larger memory and greater processing speeds.

\begin{figure}[!t]
\centering
\includegraphics[width=12cm]{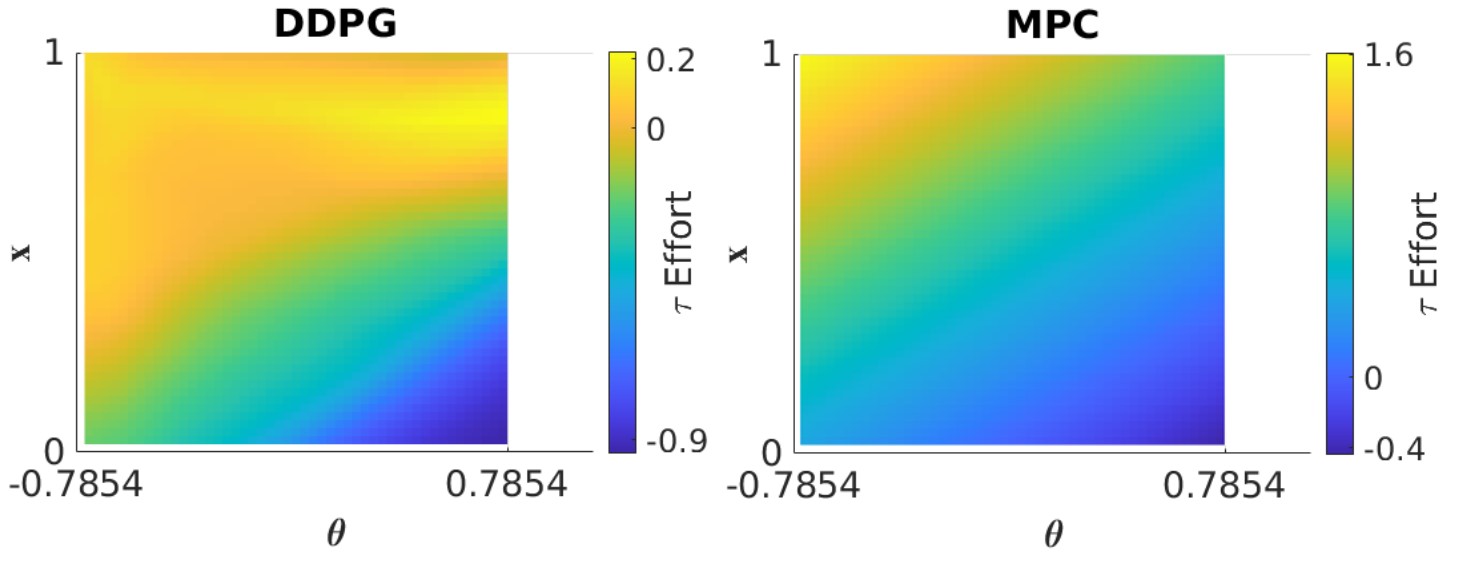}
\caption{Torque response for various $x$ vs $\theta$ scenarios}
\centering
\includegraphics[width=12cm]{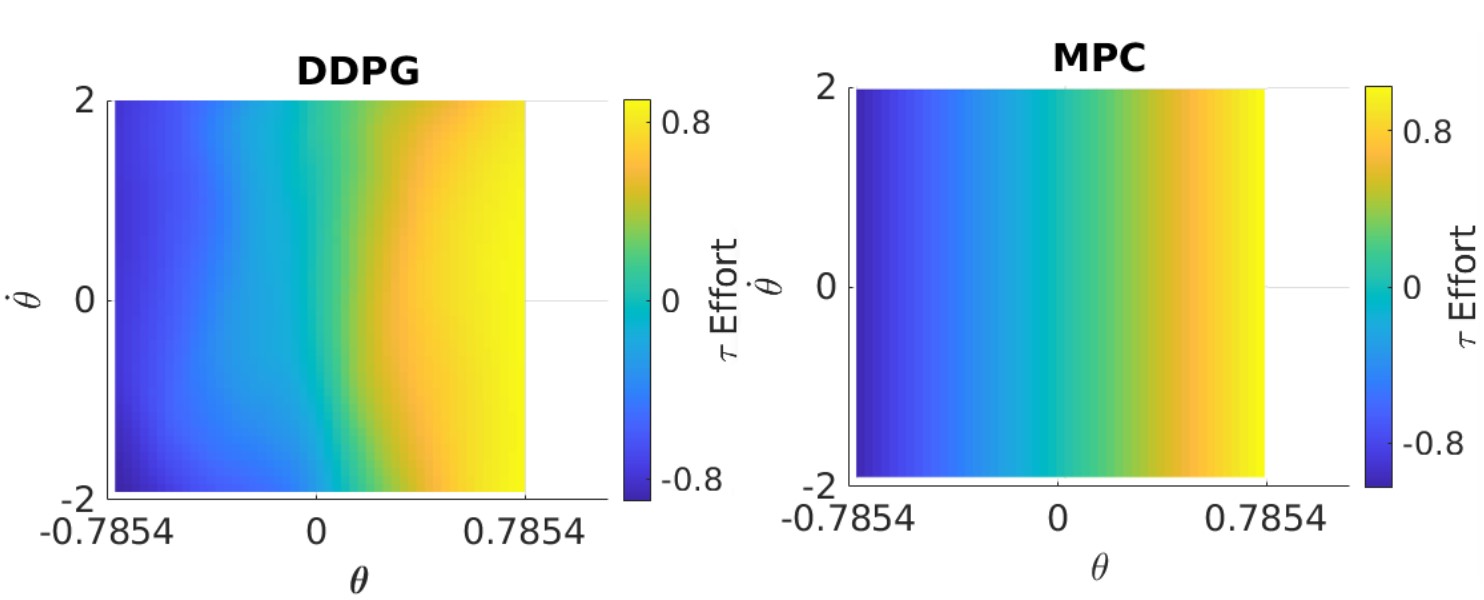}
\caption{Torque response for various $\dot{\theta}$ vs $\theta$ scenarios}
\end{figure}

The effort responses for the best performing RL controller i.e. EH based DDPG is given in Figure 10 and 11 for various combinations of $x$ vs $\theta$ and $\dot{\theta}$ vs $\theta$ initial conditions. The wheel torque effort closely resembles that of the MPC controller. It follows a similar trend in responding to various state scenarios espacially when given various angular velocity for a given angular displacement. This confirms that DDPG based RL control policy has closely converged to an MPC based predictive controller with a greater horizon.

The method of providing error history of last 6 time steps seems to be promising as it has out performed their classical counterparts. We believe this may be due to the additional information available to the policy networks about how better did the system performed as far as the trajectory is concerned. The EH based policies also reach acceptable net reward faster for this environment when compared to non EH based control policies. Here the trajectory was chosen to be some points in the x direction but this can very well be some command velocities to be followed in more comprehensive systems.

\section{Conclusion}
In this paper we discussed different RL based control algorithms, namely PPO and DDPG with and without trajectory error history inputs and implemented those on an Extendable Wheeled Inverted Pendulum (E-WIP). The system defined is a nonlinear system. To control such nonlinear systems, complicated functions which gives the required inputs to the system based on the independent state variables can be represented as Neural Networks and one can implement a pipeline where these systems learn to find an optimal control function by self-improving through reward-based learning. 
This article has compared popular Reinforcement Learning algorithms to a Model Predictive Control approach. The results indicate that RL models with error histories of the reference states perform well in balancing the robot and following the given way points. DDPGs are successful at following the trajectory exactly but PPOs struggle towards the end to stop the movement at the goal location despite more training effort.The results concluded that involving previous state errors in desired states improved the training and overall performance of the RL based controllers in the E-WIP system

The future works includes exploring a Two-wheeled Extendable approach to traverse various types of terrains and the exploration of ‘jumping’ control on the present system. Also, we would explore more robust non-linear control methods and other RL algorithms to better control more generalized trajectories.

\section{Acknowledgements}
The authors would like to express their gratitude to the Department of Applied Physics and Department of Mechanical Engineering, Delhi Technological University, New Delhi to promote research in the field of robotics and application of new AI techniques. This work would not have been completed without the support of undergraduate students in DTU Altair Laboratory, DTU, Delhi. The authors are thankful to every person involved directly or indirectly with the work and helped along the way.

\bibliographystyle{unsrt}  


\end{document}